\documentclass[a4paper, 10pt, conference]{ieeeconf}      
\usepackage{FG2021}

\FGfinalcopy 
\IEEEoverridecommandlockouts                              
\usepackage{times}
\usepackage{epsfig}
\usepackage{graphicx}
\usepackage{amsmath}
\usepackage{amssymb}
\usepackage{dblfloatfix} 
\usepackage[pagebackref=true,breaklinks=true,letterpaper=true,colorlinks,bookmarks=false]{hyperref}

\makeatletter
\newcommand{\xRightarrow}[2][]{\ext@arrow 0359\Rightarrowfill@{#1}{#2}}
\makeatother

\def\FGPaperID{****} 

\title{\LARGE \bf
FaceQgen: Semi-Supervised Deep Learning \\for Face Image Quality Assessment
}

\author{\parbox{16cm}{\centering
    {\large Javier Hernandez-Ortega, Julian Fierrez, Ignacio Serna and Aythami Morales}\\
    {\normalsize School of Engineering, Universidad Autonoma de Madrid, Spain}}
    \thanks{Support by projects PRIMA (H2020-MSCA-ITN-2019-860315), TRESPASS-ETN (H2020-MSCA-ITN-2019-860813), and BIBECA (RTI2018-101248-B-I00 MINECO/FEDER). J.H.-O. and I.S. are supported by PhD scholarships from UAM.}
}

\begin{document}

\IEEEoverridecommandlockouts\pubid{\makebox[\columnwidth]{978-1-6654-3176-7/21/\$31.00~\copyright{}2021 IEEE \hfill} \hspace{\columnsep}\makebox[\columnwidth]{ }}

\ifFGfinal
\thispagestyle{empty}
\pagestyle{empty}
\else
\author{Anonymous FG2021 submission\\ Paper ID \FGPaperID \\}
\pagestyle{plain}
\fi
\maketitle

\begin{abstract}

In this paper we develop FaceQgen\footnote{Publicly available in: https://github.com/uam-biometrics/FaceQgen}, a No-Reference Quality Assessment approach for face images based on a Generative Adversarial Network that generates a scalar quality measure related with the face recognition accuracy. FaceQgen does not require labelled quality measures for training. It is trained from scratch using the SCface database. FaceQgen applies image restoration to a face image of unknown quality, transforming it into a canonical high quality image, i.e., frontal pose, homogeneous background, etc. The quality estimation is built as the similarity between the original and the restored images, since low quality images experience bigger changes due to restoration. We compare three different numerical quality measures: a) the MSE between the original and the restored images, b) their SSIM, and c) the output score of the Discriminator of the GAN. The results demonstrate that FaceQgen's quality measures are good estimators of face recognition accuracy. Our experiments include a comparison with other quality assessment methods designed for faces and for general images, in order to position FaceQgen in the state of the art. This comparison shows that, even though FaceQgen does not surpass the best existing face quality assessment methods in terms of face recognition accuracy prediction, it achieves good enough results to demonstrate the potential of semi-supervised learning approaches for quality estimation (in particular, data-driven learning based on a single high quality image per subject), having the capacity to improve its performance in the future with adequate refinement of the model and the significant advantage over competing methods of not needing quality labels for its development. This makes FaceQgen flexible and scalable without expensive data curation.

\end{abstract}

\section{INTRODUCTION}

Nowadays most face recognition applications take place under unconstrained conditions~\cite{2018_IntelSys_Trends_Proenca}, what favors some variability factors to appear like low resolution, heterogeneous background, blur, etc., affecting severely the recognition accuracy \cite{lu2019experimental}. Consequently, it becomes necessary for recognition systems to have the ability to deal with these factors, e.g., by detecting their presence and (if possible) reducing their impact. A biometric quality measure can be defined as a function that transforms an input sample, e.g., a face image, into an estimation of its quality level~\cite{2011_QualityBio_FAlonso}. In this context, the quality of a face image can be understood as a predictor of how suitable it will be for face recognition (or other AI application), since an image with low quality will not produce reliable results when used for that application.

\begin{table*}[t]
\renewcommand{\arraystretch}{0.4}
\centering
\caption{\textbf{Comparison of relevant works in face quality assessment}, divided in traditional works and deep learning-based methods. NR = No-Reference, Acc. = Accuracy, Num. = Numerical, FS = Fully-Supervised Quality Learning, SS = Semi-Supervised Quality Learning.}


\begin{tabular}{ccccc}
\multicolumn{5}{c}{\textbf{Traditional Methods}} \\ \hline
\textit{Year, Reference}                                                              & \textit{Groundtruth} & \textit{Input} & \textit{Features}                                                                                                             & \textit{Output}                                                                                                             \\ \hline
\\
2006,~\cite{hsu2006quality}            & Human \& Acc. based         & NR, FS & Face \& image features & Num. Score \\ 
\\
2012,~\cite{abaza2012quality}            & Human based         & NR, FS & Image features & PDF \& Num. Score \\ 
\\

2012,~\cite{ferrara2012face}            & Human based         & NR, FS & $30$ ICAO compliance tests & Num. Score for each test \\ 
2019,~\cite{khodabakhsh2019subjective}           & Human \& Acc. based         & NR, FS & Face \& image features & Num. Score \\ 
\\     
\multicolumn{5}{c}{\textbf{Deep Learning Methods}} \\ \hline
\textit{Year, Reference}                                                              & \textit{Groundtruth} & \textit{Input} & \textit{Features}                                                                                                             & \textit{Output}                                                                                                             \\ \hline 
\\

2017,~\cite{zhang2017illumination}            & Human based      & NR, FS & CNN Illumination-based features & Num. Score \\ 
 \\

2018,~\cite{best2018learning}            & Human \& Acc. based         & NR, FS & CNN features & HQV \& MQV \\ 
 \\

2019,~\cite{2019_FaceQnet_Hernandez}           & Human \& Acc. based         & NR, FS & CNN features & Num. Score \\ 
\\ 
 
2020,~\cite{terhorst2020ser}           & Accuracy based         & NR, FS & CNN features & Num. Score \\ 
\\

2021,~\textbf{[FaceQgen (ours)]}           & \textbf{Human based (only high Q)}         & \textbf{NR, SS} & \textbf{Restored image}  & \textbf{Num. Score} \\
\\    
\end{tabular}

\label{table:related_works}
\end{table*}



In this paper we present FaceQgen, a No-Reference quality assessment method for face recognition based on Generative Adversarial Networks (GANs). Related methods typically use supervised learning over a training dataset labelled in the full range of targeted quality. FaceQgen is a data-driven approach based on latest deep learning methods that improves that aspect: it is trained using only one high quality sample per subject. The quality measure estimated by our method will be related to the expected accuracy of the recognition process. 

The advantage of FaceQgen is the absence of an explicit definition of quality since it is designed to learn inferring knowledge directly from high quality samples. FaceQgen does not need a specific measurement of quality for its training groundtruth and it learns to estimate the full range of face image quality in a data-driven way based on a single high quality shot per subject. Even though our driving application is face recognition, we anticipate that FaceQgen will be also useful in other applications over face images, e.g., clustering, facial features analysis, profiling, etc.


The main contributions of this work are: 1) a new quality assessment method for face images called FaceQgen designed with end-to-end learning without explicit usage of hand-crafted quality factors, which additionally doesn't need labelled quality measures for learning, and 2) we evaluate FaceQgen as a predictor for face recognition accuracy over three publicly available datasets. Our results validate the idea, and open new research opportunities to develop new quality assessment methods.

\section{Related Works}
\label{related_works}

Even though a significant number of works in face quality assessment have been proposed~\cite{schlett2021face}, to this date, it does not exist a global standard on general face quality. The closest attempts to define face quality are the technical reports published by ICAO TR 9303 and ISO/IEC 39794-5 for trying to define portrait-like images of perfect quality in order to regulate their inclusion in official documents. However, these reports do not contain the definition of a quality measure, only some guidelines for good image acquisition. A good example of standard in biometric quality measures is the case of fingerprints, where the ISO/IEC 29794-4 cites a quality measure developed by NIST, i.e., NFIQ~\cite{tabassi2004fingerprint}, to be used as quality measure. In the present work we propose a method that could be used as a face quality standard in the same way as NFIQ has been applied to fingerprints. 

In Table~\ref{table:related_works} we include a compilation of relevant works in face quality assessment. The selection has been made to be representative of the two main stages of face quality assessment research in the last $15$ years: 1) a first stage that contains works based on hand-crafted features and classic machine learning; and 2) a second stage, starting in 2017, with works powered by deep learning.


\subsection{2006-2019: Traditional Methods}

First works in face quality assessment consisted in extracting hand-crafted features from face images. Each work calculated its own quality measure using the extracted features to estimate the presence of one or several variability factors that have traditionally been considered to affect recognition accuracy, like pose, background, occlusions, illumination, and blur.

The work in~\cite{hsu2006quality} is a perfect example of this first line of research in face quality. The authors calculated several quality measures using hand-crafted algorithms, each one estimating the presence of a specific variability factor, e.g., pose, illumination, skin texture, compression artifacts, etc. Then they combined all the measures into two different global quality measures for each face image, one related to human perception and the other to recognition accuracy.


A natural evolution of hand-crafted approaches was made in~\cite{abaza2012quality} where the authors presented a global accuracy-based Face Quality Index (FQI) that was calculated by combining individual quality factors extracted from five image features: contrast, brightness, focus, sharpness, and illumination. They added synthetic effects to the original images in order to emulate real world acquisition variability. Similarly to~\cite{abaza2012quality}, the BioLab-ICAO presented in~\cite{ferrara2012face} perform $30$ individual tests checking for variability factors and returning a numerical score for each one. However, in this case the tests are designed for checking the level of compliance of face images with the ISO/ICAO standard for MRTD. Furthermore, this time the individual scores were not combined in any manner to obtain a global quality measure.




The authors of \cite{khodabakhsh2019subjective} made a study of both subjective and objective face quality measures, comparing their effect on face recognition scores. They manually labelled a face recognition database with scores related to the ease of recognizing the face. Then they compared those subjective scores with other objective scores calculated using the guidelines of ISO/IEC TR 29794-5. They found that the correlation of the subjective scores with the recognition scores outperformed the correlation of the objective scores.



Hand-crafted and traditional learning-based image processing approaches have the advantage of being interpretable since the tests are designed with the target of measuring specific image features like: blur, resolution, texture, color, etc. However, it is difficult to determine which of those features are more relevant for the task at hand. Additionally, the hand-crafted algorithms sometimes are not as accurate as expected under certain acquisition scenarios, so the results may not be totally reliable.

\subsection{2017-2021: Deep Learning Methods}



\begin{figure*}[t]
\centering
\includegraphics[width=0.9\textwidth]{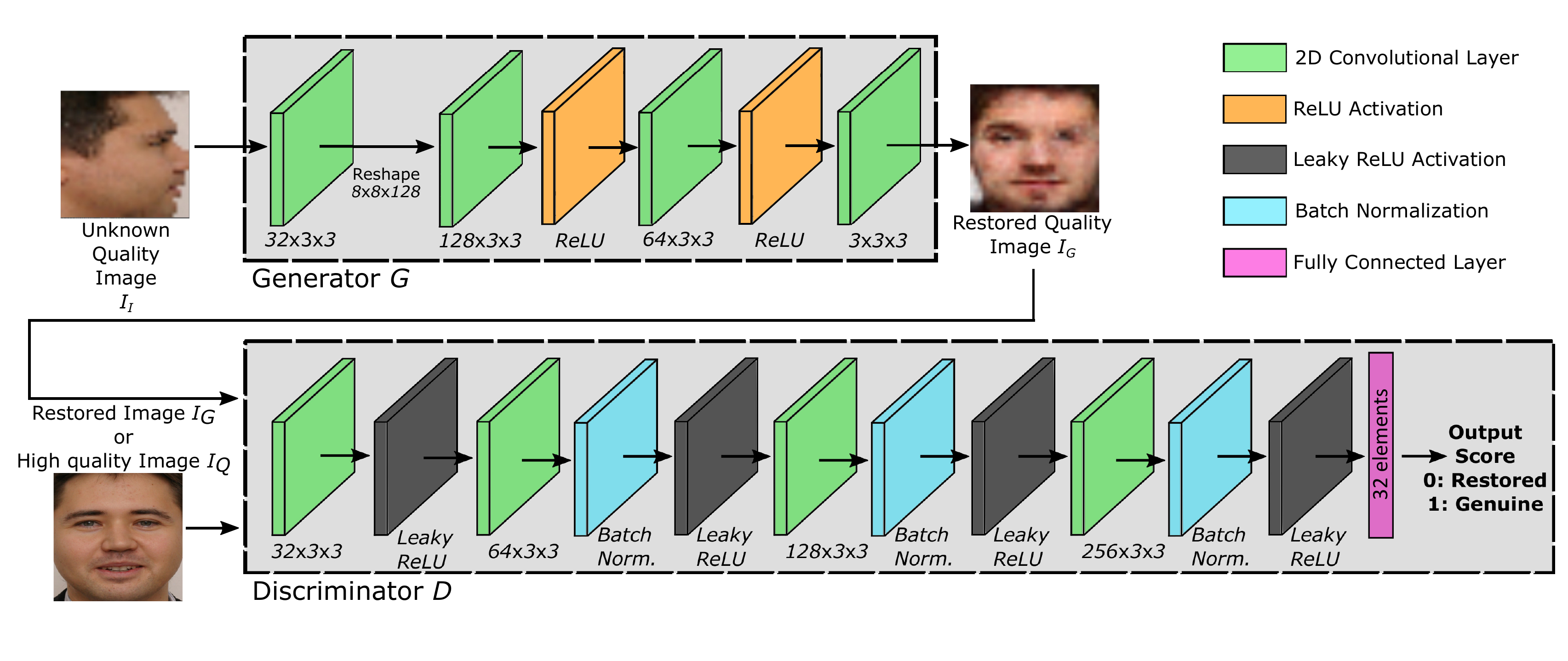} 
\caption{\textbf{FaceQgen architecture.} It comprises two different networks: i) a Generator for transforming face images into portrait-like ICAO compliant images, and ii) a Discriminator that tries to distinguish between original and restored high quality face images. For the convolutional layers ($i \times j \times k$) denotes $i$ filters with ($j \times k$) kernel size. Strides are of size $2$ for all the convolutional layers of the Generator and of size $3$ for the Discriminator. The size of the input and restored images is $32 \times 32 \times 3$ pixels.}
\label{fig:general_scheme}
\end{figure*}

Recently, inspired by the success of deep learning, face quality assessment works are also applying deep learning techniques. For example, in \cite{zhang2017illumination} the authors proposed a method based on a CNN architecture for evaluating the quality of face images based on their illumination conditions. They captured a Face Image Illumination Quality Database (FIIQD) based on human perception and they used it for training a CNN model originally designed for object classification. Another example of successful application is the work presented in~\cite{best2018learning} where the authors developed a system based on Convolutional Neural Networks (CNNs) to estimate two different quality measures, one related to human perception and other related to recognition accuracy. They found that both measures were correlated with face recognition accuracy, being HQV the most accurate of both measures.


FaceQnet~\cite{hernandez2020biometric,2019_FaceQnet_Hernandez} is a recent deep learning method based on CNNs that correlates the quality of a face image with face recognition accuracy. The authors identified the lack of data labelled with face quality information and decided to use the BioLab-ICAO software for labelling training data automatically with quality labels related to ICAO compliance. Then they used that groundtruth to fine-tune a CNN model pretrained for face recognition, adapting it to face quality estimation. FaceQnet's quality measures demonstrated to be highly correlated with face recognition accuracy, even when testing them against commercial and open source face recognisers not seen during the training process. 

In ~\cite{terhorst2020ser} the authors proposed another deep learning method for correlating face image quality with recognition accuracy, but this time it consisted in an unsupervised face quality assessment method that extracted face embeddings from CNNs pretrained for face recognition. Images with high quality should present robustness through its different embeddings while low quality images are expected to vary significantly. This hypothesis was also used in~\cite{terhorst2021comprehensive}, where the authors wanted to predict $74$ attributes from face templates. They used a quality measure for discarding the $50$\% of images with lower quality, since their embeddings variations made it impossible to predict attributes from them.

Works like~\cite{2019_FaceQnet_Hernandez,hernandez2020biometric,terhorst2020ser} took part in the FRVT-QA organised by NIST~\cite{DraftREPORTnistFRVTQ}, obtaining results that have set the state of the art in face quality assessment, showing the high correlation between the estimated quality values and the actual face recognition accuracy of commercial face matchers.

Additionally to the mentioned regular CNN architectures, Generative Adversarial Networks have shown to be powerful solutions for tasks in which generating new data is a need, e.g., image restoration. The method we propose in this paper, FaceQgen, is inspired in works like~\cite{ma2020active} which explored GAN architectures for general image quality assessment. Our model transforms a face image into a canonical high quality face image: frontal pose, homogeneous background, etc, and then we use the original and the restored image to estimate a face quality measure.

Most of the related works in the literature, like~\cite{best2018learning} and~\cite{hernandez2020biometric}, belong to the category of Fully-Supervised methods. Nevertheless, the scarcity of correctly labelled data is the main issue of these methods, not only due to the high amount of data needed to train them, but also to the difficulty of labelling data with explicit quality values without making errors or introducing human bias~\cite{terhorst2021comprehensive}. Semi-Supervised methods emerged for dealing with those limitations as they do not need a full set of training labels for learning. 

For FaceQgen we adopted a Semi-Supervised approach. In order to train our model we just need to select one high quality image (as near as possible to ICAO compliance) for each subject in the training database, without giving any numerical quality label to it. The remaining images of each subject will serve as training samples of unknown quality. Those training samples will ideally range from images of very low quality to other images near to ICAO compliance, containing several sources of variability that affect quality, e.g., blur, non-frontal pose, low resolution, etc. Building and curating a training database of this type is much cheaper, precise, and less prone to labelling bias than labelling all the images with numerical quality values, as it is required in all existing Fully-Supervised face quality methods (see Table~\ref{table:related_works}).






\section{FaceQgen Design}


The ICAO TR 9303 describes the desired conditions for capturing face images of high quality for its use in electronic documents like passports or IDs. The presence of factors such as low resolution, bad illumination, extreme pose, or blur, will determine the quality of a face image. In this work we define the quality of an image $A$ as a combination of the presence of those different factors following the next equation:

\pagebreak



\begin{equation*}
    Q(A) = \sum Ind(A)\\ 
\end{equation*}
\begin{equation}\label{ecuacion1}
\begin{split} 
   \sum Ind(A) &= Resolution(A) + Pose(A) + ...\\
...\ &Illumination(A) + Other Factors(A)
\end{split}
\end{equation}

\noindent where the indicators of quality $Ind(A)$ are defined as the individual measures of compliance with each one of the quality requirements described by ICAO. For example, if $A$ has high resolution, its indicator $Resolution(A)$ will be elevated. 

Based on the definition of Quality $Q$ set in (\ref{ecuacion1}), we made the next assumptions in order to obtain the quality groundtruth to train FaceQgen:

\begin{itemize}
    \item Two face images $A$ and $B$, both from the same person and of high quality, i.e., frontal pose, high resolution, good illumination, etc., will always be very similar (low intra-subject variability). In these cases, the similarity score $S$ obtained when comparing the two images will be high:
    \begin{equation}\label{ecuacion2}
       \text{If}\ Q(A)\cong Q(B)\ \&\ \text{high}\ Q(A)\ \Rightarrow \text{high}\ S(A,B)\\
    \end{equation}
    
    \item On the contrary, if one of the two images being compared is of uncertain quality, the mated similarity score $S$ between them will heavily depend of the differences in the quality factors of both images.

\end{itemize}

In order to obtain a quality groundtruth based on similarity scores, what we need to prove is if the relationship between $Q$ and $S$ stated in (\ref{ecuacion2}) is also true in the opposite direction, i.e., if the similarity score $S$ between an image $A$ of perfect quality and a picture $B$ of unknown quality can be taken as an accurate estimation of $Q(B)$:
\begin{equation}\label{ecuacion3}
\text{If}\ \text{high}\ S(A,B)\ \&\ \text{high}\ Q(A)\ \xRightarrow{\mathit{?}} Q(A)\cong Q(B) 
\end{equation}

To demonstrate the validity of (\ref{ecuacion3}) let's suppose a case in which we know that the similarity score between $A$ and $B$ has a high value, and that $Q(A) \neq Q(B)$ (high intra-subject variability). This assumption describes an impossible situation due to the very definition of mated similarity scores which must depend only of the intra-subject variability. Therefore, if we are sure that $Q(A)$ is high, the only way that the similarity score between $A$ and $B$ can be high is when $Q(B)$ is also elevated. This fact shows the validity of (\ref{ecuacion3}) and enables us to obtain a machine-generated quality groundtruth for an image $B$ of unknown quality just using its mated similarity score with an image $A$ of known high quality.

Based on (\ref{ecuacion3}), we trained FaceQgen to perform face image restoration, i.e., transforming an input face image $I_I$ into a high quality version of itself $I_G$ (close to ICAO compliance). If $I_I$ is of low quality, then the differences between $I_I$ and $I_G$ will be large. Otherwise, $I_I$ and $I_G$ will be similar. Then the quality of the original image will be estimated by measuring the similarity between both images. The lower the similarity, the lower the quality measure. It must be noted that we only consider mated scores for the definition of quality following the guidelines of NIST in its FRVT-QA~\cite{DraftREPORTnistFRVTQ}. Non-mated comparison scores do not depend so heavily of the quality of the samples, but mainly of the differences between the features of the two face images, which change significantly between different identities.

Fig.~\ref{fig:general_scheme} shows the architecture of FaceQgen. The GAN is composed of two different CNNs learning in an adversarial setup: a Generator $G$ and a Discriminator $D$. FaceQgen is designed as follows. For each face image of unknown quality $I_I$, $G$ will try to restore it obtaining a high quality and distortion-free face image $I_G$. If the training dataset contains a high number of images with several samples of diverse variations (e.g., blur, lateral pose, low resolution, occlusions, etc.) $G$ will learn how to restore all kind of variations that may appear in the test images without explicit modeling or analysis of individual covariates.


The target of $D$ consists in distinguishing between restored images and images of genuine high quality to improve the accuracy of $G$. For that purpose, the output of $D$ will be a numerical score between $0$ (the Discriminator predicts that the image was generated by $G$), and $1$ ($D$ predicts that the image presents genuine high quality). The GAN learning process optimizes iteratively $G$ in a data-driven way until $D$, also optimized iteratively, is not able to distinguish between restored and genuine high quality.  

\subsection{FaceQgen: Networks}

We decided to adopt a VGG-like architecture for $G$ \cite{simonyan2014very} as it is a well-known approach that has shown its effectiveness in computer vision applications like image classification and regression. Details of the network structure can be seen in Fig.~\ref{fig:general_scheme}. The Generator $G$ receives RGB color face images (detected and cropped) with a shape of $32 \times 32 \times 3$ and with their values normalized to the $[0,1]$ range for improving the stability of the network. The final output of $G$ is another $32 \times 32 \times 3$ RGB image that is expected to be a high quality restored version of the input image. When using small images some of the high frequency details can be lost, but based on our empirical experience, a size of $32 \times 32$ presents a good balance between the details in the image and the amount of parameters of the Generator model. With more training data available, it would be possible to increase the size of the input images to preserve fine details.

Fig~\ref{fig:general_scheme} also shows that for $D$ we again adopted a VGG-like architecture. This time the convolutional layers are followed by Leaky ReLU activations and batch normalization to keep the output values inside the desired range and improve the learning process. We have followed the last convolutional layer with a Fully Connected (FC) layer and an output layer that computes a classification score between \emph{Genuine} high quality images ($1$) and \emph{Restored} high quality images ($0$).

\subsection{FaceQgen: Training Protocol}

\begin{figure*}[t]
\centering
\includegraphics[width=0.9\linewidth]{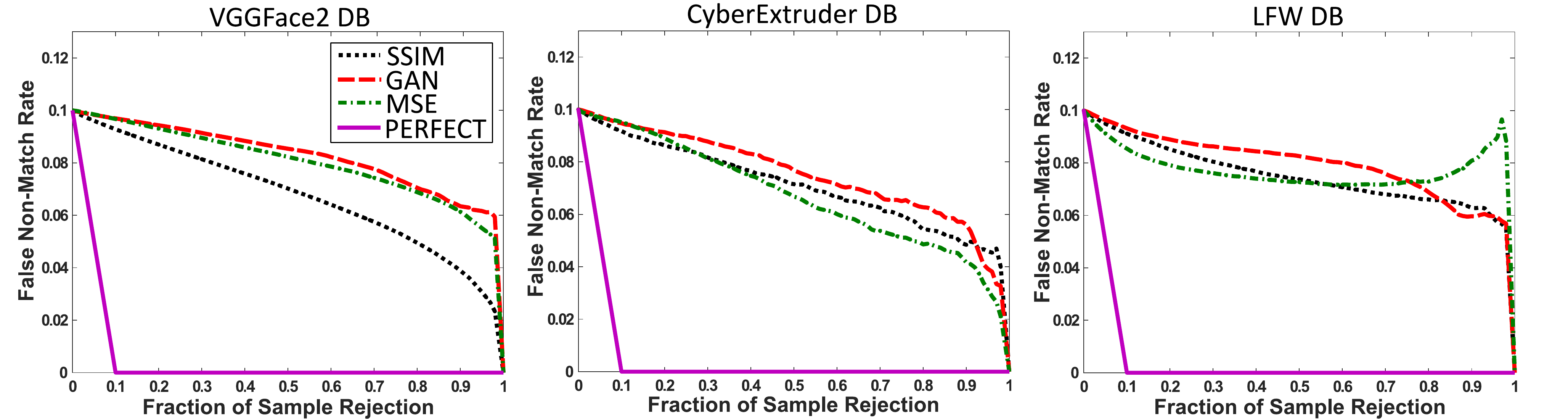} 
\caption{\textbf{Error versus Reject Curves for the three test datasets.} Verification scores were obtained using FaceNet. Three different quality measures have been evaluated: Mean Squared Error (MSE), Structural Similarity (SSIM), and the output score from our GAN Discriminator. The initial FNMR has been set to $10$\%. The PERFECT curve was calculated using $\max(\textrm{FNMR} - \textrm{Fraction of Sample Rejection} , 0)$. The closer the curve of a quality measure is to the PERFECT line, the more accurate that quality measure will be estimating face recognition accuracy. (Color image.)}
\label{fig:ERC_curves}
\end{figure*}




We decided to train FaceQgen using \textbf{SCface} \cite{grgic2011scface}, which contains $4$,$160$ images (in infrared and visible spectrum) of $130$ subjects. The quality of the images comprise a wide range, going from images captured using surveillance cameras to portrait-like images. We selected a portrait image per subject as groundtruth and all the remaining images of each subject as input samples for the learning process, which consisted in two different stages. 


First, we trained the Generator alone to have a good initialization of its weights. We trained $G$ for $10$,$000$ iterations over the entire dataset with a batch size of $64$ images using Adam optimizer (initial learning rate of $0.001$). 

Since face images are highly structured and their pixels have strong spatial dependencies with the values of close pixels likely to be quite similar, we decided to use a loss function for the Generator different than the common Mean Squared Error (MSE). MSE only takes into account the mean difference between the pixels of both images but without considering other properties like the structure of the pixels. Considering those properties is desirable for this specific application in which we are trying to obtain images capable of passing as real faces. With that target in mind, we adopted the Structural Similarity Index Measure (SSIM) described in~\cite{wang2004image} as the function loss of $G$. SSIM compares patterns of pixel intensities that have been normalized for luminance and contrast. Its authors decided to separate the influence of the illumination and contrast from the structural information of the image since they are relatively independent. Thus, SSIM separates the similarity metric into three different measures, i.e., luminance, contrast, and structure, that are subsequently combined in a final value. Using SSIM, the training of the Generator will be focused to obtain restored face images more similar to real faces of high quality.



Secondly, after initializing $G$ correctly, we train the Generator and the Discriminator together for another $10$,$000$ iterations. For each iteration, we first train the Discriminator separately along the entire dataset. Then, we clip the weights of $D$ into the $[-0.05,0.05]$ range (decided empirically) for improving the convergence of the training process. Finally, we froze all the weights of $D$ and we train the complete GAN on the entire dataset. In both cases the batch size and the optimizers are the same than in the first training stage (when we trained $G$ alone). For this training stage we adopted binary cross-entropy as the loss function. After this training process, FaceQgen becomes capable of restoring face images of unknown quality. Finally, we chose to measure the distance (or similarity) between $I_I$ and $I_G$ in three different manners: 1) their MSE, 2) their SSIM, and 3) the output score from $D$ when it is fed with $I_G$.

As we stated previously, MSE is an error measure that does not consider the structure of the pixels in the image, so even when $I_I$ and $I_G$ get a low MSE when being compared, the restored image could be easily detected as a fake face by an external observer (or the Discriminator $D$) due to errors in the structure of the restored face. On the contrary, a restored face image $I_G$ with a realistic structure could easily spoof an external viewer, even when obtaining a high MSE in its comparison with $I_I$, likely caused by differences in color and luminance, factors that may not have as much impact in face recognition as the facial structure. By comparing MSE and SSIM we will compare the performance of a similarity measure that only takes into account the absolute errors between the two images (MSE) to another measure that also takes into account the structure of both images (SSIM). If an image $I_I$ is of good quality, we can expect its restored version $I_G$ to be similar in structure, even when $G$ may change its color or luminance due to the limitations in its training process.

Finally, we decided to use the output score of the Discriminator as our third quality score, as it will give us an estimation of the restoration capabilities of $G$. $D$ has been trained to distinguish between restored images and original images of high quality, while $G$ has been trained for trying to fool $D$ with restored images. If the restoration capabilities of $G$ are accurate, the restored images $I_G$ will be indistinguishable to images of genuine high quality for the Discriminator, and its output score will not be a good quality estimator. On the contrary, if $G$ is not capable of restoring low quality images, $D$ will distinguish easily between restored images and high quality images, and the output score of $D$ will be a good quality estimator.


\section{FaceQgen Evaluation}

\begin{figure*}[t]
\centering
\includegraphics[width=0.93\linewidth]{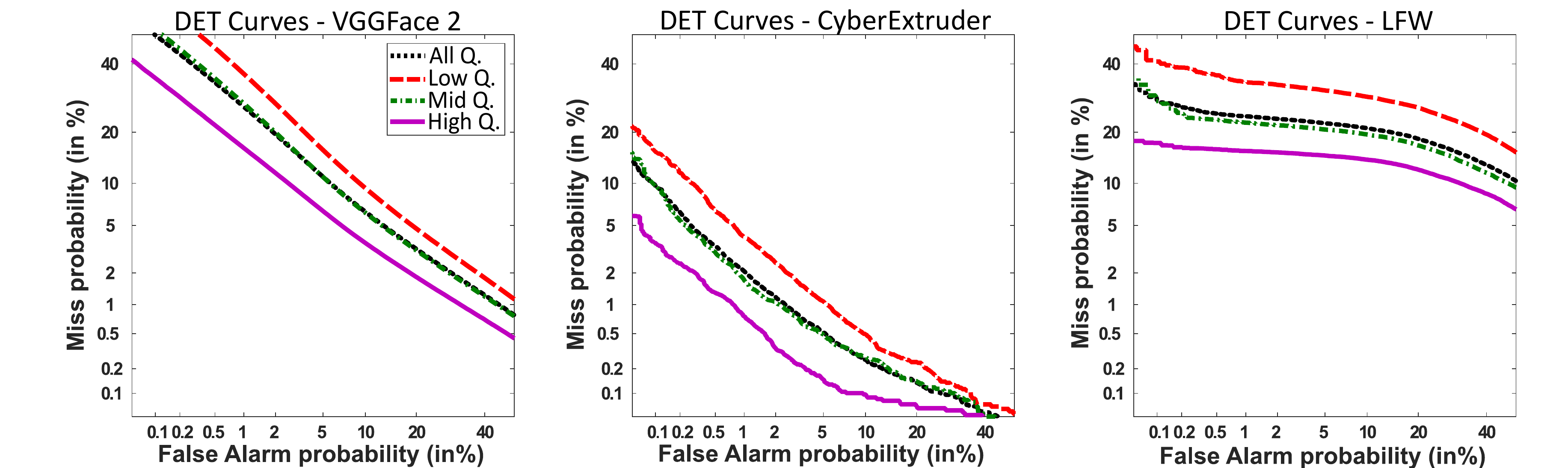} 
\caption{\textbf{DET Curves for the three test databases.} Verification scores were obtained using FaceNet. Images have been divided in three different FaceQgen + SSIM based quality ranges: low, medium, and high quality. As expected False Alarm probability and Miss probability decrease when the mean quality of the images increases.}
\label{fig:DET_curves}
\end{figure*}


We tested FaceQgen on $3$ different datasets whose images present diverse levels of quality: 1) \textbf{VGGFace2} \cite{cao2018vggface2} presents unconstrained acquisition conditions with variations in pose, illumination, background, etc.; 2) \textbf{CyberExtruder\footnote{The Ultimate Face data set was provided by CyberExtruder.com, Inc. 1401 Valley Road, Wayne, New Jersey, 07470, USA}} with images downloaded from Internet, also with diverse acquisition conditions; and 3) \textbf{LFW} \cite{LFWTech} that was designed for studying the problem of unconstrained face recognition.

We used all the images in each one of the three evaluation databases. First, we detected the face in each image using MTCNN as our face detector~\cite{zhang2016joint}. Then we cropped and resized each face to $32 \times 32 \times 3$ pixels, obtaining $I_I$ to feed $G$. The output of the generator is the restored face image $I_G$. 
Then, using both $I_I$ and $I_G$ we computed the three different quality values: MSE, SSIM, and $D$ score.

Fig.~\ref{fig:ERC_curves} shows Error versus Reject Curves (ERC)~\cite{grother2007performance,DraftREPORTnistFRVTQ} comparing the accuracy of the three proposed quality measures to estimate face recognition accuracy. First, we fixed the initial verification threshold to obtain a FNMR of $10$\% using all the mated pairs indistinctly. Then we began to discard the images with the lowest quality measures obtaining new values of FNMR that should decrease each time since the images with the worst quality were being rejected. 

An ideal quality measure would present a perfect correlation with recognition accuracy, which would entail that its associated ERC curve would coincide with the FNMR of a given recognition system. In the ERC plots shown in Fig.~\ref{fig:ERC_curves} that perfect correlation is represented by the curves designed as PERFECT.  Therefore, the closer a ERC to the PERFECT curve, the more accurate the quality measure will be.

\begin{figure*}[t]
\centering
\includegraphics[width=0.93\linewidth]{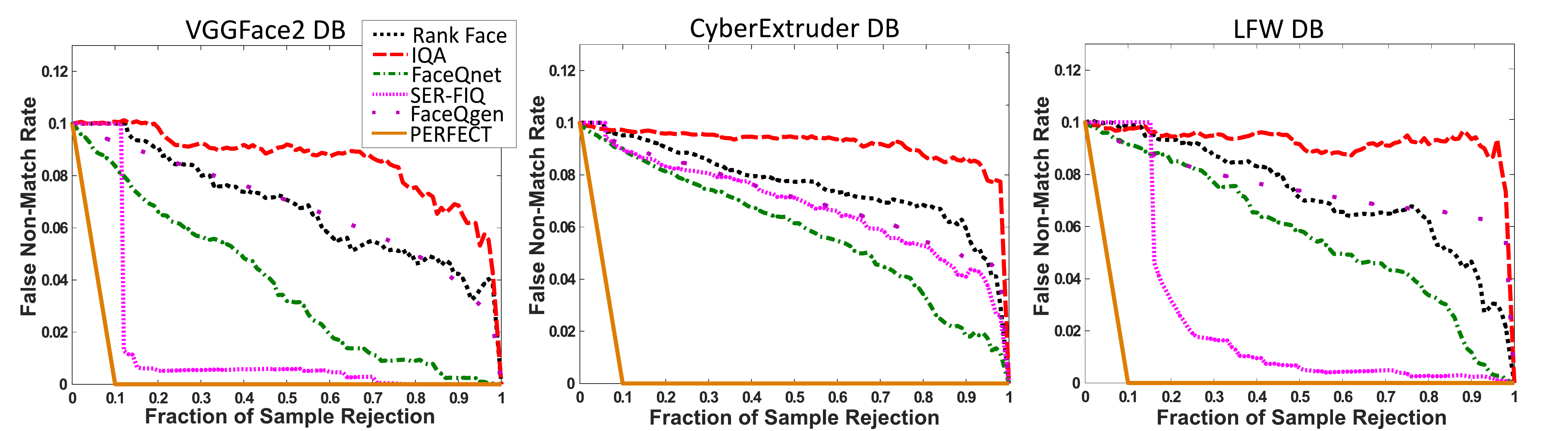} 
\caption{\textbf{Error versus Reject Curves for the three test datasets} obtained with the FaceNet comparator for $5$ state-of-the-art algorithms. The initial FNMR has been set to $10$\%. Fractions of the images with lowest quality measures have been removed consecutively. Five different QA algorithms have been used for
obtaining quality measures of the testing images: a general Image Quality Assessment (IQA) method \cite{idealods2018imagequalityassessment}, a method for face QA based on handcrafted features (Rank Face IQA) \cite{chen2014face}, FaceQnet \cite{hernandez2020biometric,2019_FaceQnet_Hernandez}, SER-FIQ ~\cite{terhorst2021comprehensive}, and the method proposed in this paper: FaceQgen. The line labelled PERFECT is generated using $\max(\textrm{FNMR} - \textrm{Fraction of Sample Rejection} , 0)$. The closer the quality algorithm line is to the PERFECT line, the more related the quality measure is to face recognition accuracy. (Color image.)}
\label{fig:ERC_curves2}
\end{figure*}

FaceNet~\cite{schroff2015facenet} is the face recognizer we used to calculate the ERC associated to each combination of database and quality measure. Analysing the ERCs in Fig.~\ref{fig:ERC_curves}, the combination of FaceQgen and SSIM showed the most robust performance through the different datasets, being even the most accurate option when evaluating on VGGFace2, i.e., the database that contains the highest number of images and the most diverse levels of quality. This observation makes reasonable to conclude that the SSIM-based quality measure will be the most robust option when used in diverse scenarios. Predictably, MSE obtained less accurate results than SSIM since it does not have into account the spatial distribution of pixels in the images. 

For a face image to be considered of high quality, having a regular shape and structure, e.g., having the nose in the zone center of the bounding box of the face, is more relevant than presenting a specific color range or a particular level of luminance. In the case of face images $I_I$ of really low quality, for example those with lateral poses, some facial parts like the nose, the ears, or the mouth will not be in the same zone than they were expected to be in a frontal face image of high quality. This type of low quality images may obtain a low MSE in those cases in which $I_I$ and $I_G$ share other properties like their color and luminance, being therefore misclassified as high quality face images. On the contrary, for those cases SSIM is managing to capture the differences in the structure of these types of images, making FaceQgen + SSIM a more accurate option for the final quality measure than FaceQgen + MSE.


Finally, $D$ quality scores also show correlation with the accuracy of face recognition, but not so accurate as the two other measures. This means that the Generator $G$, that has been trained to create restored images capable to fool the Discriminator, is not always being capable of restoring all low quality images to make them indistinguishable from images of genuine high quality. If $G$ was always restoring the input images perfectly, the output of $D$ would always be perfect and the FNMR should vary randomly when discarding fractions of images. However, Fig.~\ref{fig:ERC_curves} shows that the FNMR decreases when discarding the images with the lowest $D$ scores. We think this is because the test databases contain images captured completely ``in the wild'' that have variability factors never seen during the training process of the GAN. 

These results confirm us that $G$ does not work equally well on all input images. We will need a higher amount of training data with more variability factors to train a more robust GAN for face quality estimation. Our current training database (SCFace) is only composed by images of students, professors, and employees at University of Zagreb. Gender is unbalanced as $114$ of the $130$ subjects are males while only $16$ are females, and skin color is also unbalanced as all participants are Caucasians. These numbers make clear that $G$ will not perform well when restoring images of females or non-white people~\cite{terhorst2021comprehensive}.

For our second experiment we decided to obtain additional results using the combination of FaceQgen + SSIM since it showed to be the most robust quality measure. We took all the images from the testing datasets and we divided them in three quality ranges each one with the same number of samples: the third of the images with the lowest quality measures (Low Quality), the third of the images with medium quality measures (Medium Quality), and finally the third with the highest quality scores (High Quality). Again, the recognizer used to obtain the verification scores was FaceNet. Fig.~\ref{fig:DET_curves} shows the DET curves for each combination of dataset and quality range. As expected, the error rates generally decrease with the growth of the mean quality of the samples, showing that the correlation between quality and face recognition accuracy is strong even when evaluating on different datasets than the one used for training what demonstrates the potential of FaceQgen as predictor of face quality related to face recognition accuracy.

\subsection{Comparison to the State of the Art}
\label{comparison_state}

In the last experiment of this evaluation we compared the accuracy of our best quality measure, i.e., FaceQgen + SSIM, against other Image Quality Assessment (IQA) methods. We decided to compare FaceQgen to Rank Face IQA, a Face IQA method for face recognition based on hand-crafted features~\cite{chen2014face}. We also implemented a method designed for general IQA~\cite{idealods2018imagequalityassessment} to check how well a general IQA algorithm performs when applied to face images. Additionally, we included SER-FIQ \cite{terhorst2020ser} and FaceQnet~\cite{hernandez2020biometric,2019_FaceQnet_Hernandez} in the comparison as representatives of state-of-the-art deep learning Face IQA methods. We computed an ERC for each combination of one testing database and one face quality measure. 

The ERC plots in Fig.~\ref{fig:ERC_curves2} show that SER-FIQ obtains the best global results for VGGFace2 and LFW databases when discarding images of low quality, while FaceQnet was the quality measure that obtained the best results for the CyberExtruder database (the closest to the PERFECT curve). The method proposed in this paper, FaceQgen, was always in a middle level of performance, neither the best nor the worst quality measure. FaceQgen + SSIM did not achieve so good global results compared to state-of-the-art deep learning methods but it still obtained a significantly good performance when discarding images of low quality, with its scores showing a good level of correlation with face recognition accuracy similar to the ones of Rank Face IQA and higher than the ones obtained with the general IQA measure. 

As expected, the general IQA measure performed the worst since it was designed for detecting quality factors of generic images without considering recognition performance. Therefore it can detect quality factors located in regions of the image outside the face that will not affect the accuracy of face recognition (after successful face detection). Finally, the hand-crafted method from~\cite{chen2014face} obtained good results but it might perform worse when facing data from other databases and/or scenarios since it can be difficult to adjust to datasets with other types of images and variability factors. Conversely, deep learning alternatives like FaceQgen have the potential to be easily adjustable to any possible scenario through fine-tuning.

After analysing all the ERC plots, it can be stated that FaceQgen generates quality measures more correlated with the accuracy of face recognition compared to general IQA and hand-crafted Face IQA methods, but less correlated compared to QA methods based on deep learning (FaceQnet and SER-FIQ). However, it must be remarked that this version of FaceQgen is a proof of concept designed to show the viability of this Semi-Supervised approach based on GANs, which has the big advantage of not requiring to label the training database with numerical quality measures (it just needs one high quality training image for each subject). This makes FaceQgen flexible and easily scalable without expensive data curation not like competing methods.


\section{Conclusion}

We developed and evaluated FaceQgen\footnote{Publicly available in: https://github.com/uam-biometrics/FaceQgen}, a face quality assessment method based on image restoration using GANs. Our method is capable of inferring biometric quality directly from face images. We compared three different similarity measures between the original and the restored images: SSIM, MSE, and the output of the Discriminator of FaceQgen. Faces of higher biometric quality experience less transformations during restoration, so their similarity values will be higher than the ones from lower quality images. 

In our experiments we extracted quality measures with FaceQgen for $3$ test databases: VGGFace2, CyberExtruder, and LFW. Then we evaluated the quality measures using ERCs and we identified FaceQgen + SSIM as the most robust option. The results also demonstrated that the Generator is not always capable of restoring the input images due to the limitations of the training database, e.g., never seen low quality factors or gender and skin-tone under-representation.

After that, we used the FaceQgen + SSIM option to divide the images from the $3$ test databases into $3$ different quality ranges according to their quality measures: low, medium, and high, and we measured face recognition accuracy. FaceQgen showed to be a reliable face quality measure related to face recognition accuracy as we always achieved a higher accuracy using only images of the high quality range compared with using all the images regardless of their quality.

Finally, we compared our most accurate face quality measure (FaceQgen + SSIM) to other quality assessment methods designed for faces and for general images. FaceQgen performed in a middle level of accuracy, but having into account that this version was conceived as a proof of concept that has room for improvement, the results validated the idea of training end-to-end quality assessment without the need of explicit numerical quality labels. 

As future improvements, training FaceQgen with a larger database could make the restoration process more precise. Exploring other architectures can also be beneficial for better image restoration, e.g., variational autoencoders. The accuracy of FaceQgen's quality measures should be tested with other state-of-the-art face recognizers as well, e.g., ArcFace, to check how well they generalize.

Finally, for future work we also suggest to investigate: how the proposed GAN-based methods can contribute to disentangle individual quality factors~\cite{FaceQvec} (as required in related work around biometric quality, e.g., ISO/IEC WD 29794-5); applications of quality assessment such as biometric presentation attack detection~\cite{2019_BookPAD2_IntroFacePAD_JHO}; and studying/removing biases that can affect our algorithms across different populations~\cite{serna2021ifbid}.





{\small
\bibliographystyle{ieee}
\bibliography{egbib}
}

\end{document}